\documentclass[conference]{IEEEtran}
\IEEEoverridecommandlockouts
\usepackage{amsmath,amssymb,amsfonts}
\usepackage{algorithmic}
\usepackage{graphicx}
\usepackage{textcomp}
\usepackage{xcolor}
\usepackage[backend=biber,style=numeric,sorting=ynt]{biblatex}

\begin{document}

\title{Predicting Temperature of Major Cities Using Machine Learning and Deep Learning}

\author{Wasiou Jaharabi\textsuperscript{a,*}, MD Ibrahim Al Hossain\textsuperscript{a}, Rownak Tahmid\textsuperscript{a},\\
Md. Zuhayer Islam\textsuperscript{a}, T.M. Saad Rayhan\textsuperscript{a}, Arif Shakil\textsuperscript{b}
\\

\textsuperscript{a}Depertment of Computer Science and Engineering, BRAC University, Dhaka, Bangladesh\\

\textsuperscript{*}Corresponding author; email: wasiou.jaharabi@g.bracu.ac.bd}
\maketitle

\begin{abstract}
Currently, the issue that concerns the world leaders most is climate change for its effect on agriculture, environment and economies of daily life. So, to combat this, temperature prediction with strong accuracy is vital. So far, the most effective widely used measure for such forecasting is Numerical weather prediction (NWP) which is a mathematical model that needs broad data from different applications to make predictions. This expensive, time and labor consuming work can be minimized through making such predictions using Machine learning algorithms. Using the database made by University of Dayton which consists the change of temperature in major cities we used the Time Series Analysis method where we use LSTM for the purpose of turning existing data into a tool for future prediction. LSTM takes the long-term data as well as any short-term exceptions or anomalies that may have occurred and calculates trend, seasonality and the stationarity of a data. By using models such as ARIMA, SARIMA, Prophet with the concept of RNN and LSTM we can, filter out any abnormalities, preprocess the data compare it with previous trends and make a prediction of future trends. Also, seasonality and stationarity help us analyze the reoccurrence or repeat over one year variable and removes the constrain of time in which the data was dependent so see the general changes that are predicted. By doing so we managed to make prediction of the temperature of different cities during any time in future based on available data and built a method of accurate prediction. This document contains our methodology for being able to make such predictions.
\end{abstract}


\section{Introduction}

This research was done with an aim to predict future temperatures using machine learning and deep learning. Rising temperature has been one of the largest problems in recent times so the research we have done below was done with the aim of developing a highly accurate temperature prediction method which will enable us to predict future temperature of major cities of the world which can be used to study the rise of temperature better. Moreover, forecasting temperature is dependent on a lot of data. These datasets consist data based on certain regions and the trend of the change of temperature of those regions. It will provide data based on the geography of the region, regional activities, regional timeline etc. Such large database is difficult to collect and can only predict up to a near future. But, in order to build a system capable of not just short-term but also long-term predictions based on the previous temperature data available we need the help of machine learning and deep learning approach. These work with time series. Sometimes we can see that in the prediction there are some errors in the graph. Sometimes the errors are too high. So, what is really done in this case is some error calculation is used to get the exact forecast temperature. Errors can be corrected by viewing them as amplitude errors. It is a very nonlinear optimization problem. \\
In order to do so we used machine learning and deep learning to develop an accurate prediction method using a database of monthly temperatures of major cities. The temperature graph is a non linear graph. In most cases, ML is used but in time series analysis Artificial Neural Networks(ANN) and Support Vector Machines (SVM) is used. MultiLayer Perceptron Neural Networks (MLPNN) and Radial Basis Function Neural Networks (RBFNN) developed the temperature prediction. These are the parts of ANN. Levenberg–Marquardt and Gradient Descent Is mostly used to optimize the algorithms. Deep learning is used to make predictions accurately. To forecast hourly air temperature people use Long Short Term Memory (LSTM) Recurrent Neural Networks (RNN). By running time series analysis using LSTM and other methods we constructed our prediction models. \\ 
\subsection{Motivation}
Looking at the impact of the climate change and the future consequences we will have to face regarding this matter we were motivated based on how we can participate in research done on climate change and it’s effect on temperature and make future predictions on which works can be done to tackle climate change. The threat of climate change upon mankind is now greater than ever. One of if not the most devastating consequences of climate change is the constant rise of global temperature. This trend of rise of temperature has been present and constant since the beginning of industrial revolution up to the recent decades. It has been attributed to many natural calamities such as rising sea level, drought, change in weather, harvest failure, extension of many species and so on. In order to take counteractive measurements against this problem regarding rising temperature it is important we obtain a better understanding of this situation. One of the better ways of doing so is to study through the available data we have on temperature and work through them. Also, to understand the severity of our current situation and to paint a picture of the future based on current trends we should find a viable way of predicting temperature. Such predictions will paint a clear picture of what other problems we will face in the future and the time we have left to make significant changes. It will also give us information that will come handy when measuring the effectiveness and viability of the actions taken against rising temperature. \\

\subsection{Research problem}
Working with data on temperature has some challenges to face. Anything that is related to the environment is highly unpredictable. Many natural causes or otherwise can result in drastic ups and downs in data. Moreover, it is also difficult to ensure the accuracy of the data. Different methods of data collections can lead to different results sometimes so it needs to be ensured the accuracy is maintained. But using highly accurate dataset sometimes limits the data as the modern approaches of data collection only can give data of very recent times. Which means, going further back in time the data are most of the time collected through orthodox methods where the accuracy can be questioned.\\

Also another challenge we have to face is in case of temperature the changes are often slow and it takes a while before any significant change becomes visible. The problem it creates is if the change between two base points is too far negligible and the graph becomes somewhat constant it is difficult to imply time series analysis and get results. Because, in time series the values need to be stationary for this method to work. Because in time series if something has a particular behavior pattern it follows over time it will assume that it will happen the same in the future and mitigate other possibilities. So, to ensure our data is stationary we need to take our bases from the data in a way that they maintain a constant mean, a constant variance and autocovariance that does not depend on time. At the same time the data should also maintain an accurate average to preserve overall accuracy.\\

Finally, we need to ensure that the trend and seasonality of the data is well preserved. All over the world. There are cities with short summers and long winters and vice versa. So, the trend and seasonality should come differently based on latitude, longitude and the location on the axis etc. Which means we can’t predict all the locations the same way and will need to take other significant variables into measure as well. \\

\subsection{Research objective}
The objective of this research is to create a model for predicting temperature changes which will help us to obtain better understanding in regards to climate change. Rise of temperature has been a matter of great concern all over the world. It has been identified as the root cause of many natural disasters and other problems. Major cities all over the world are concerned about bringing significant changes and controlling this trend of temperature rise. But without an insight into the future it's difficult to assess our current condition and set any benchmark. Our goal is to assist in making those benchmarks easier with predictions. Also with vast information that can be gathered from our model such as trends in changing or the seasonality of temperature rise and fall will give environmental researchers a lot of insight on the matter and drastically boost their progress. \\
\section{Background}
\subsection{Time Series}
For this project the most prominent algorithm is going to be Time Series Analysis. In this model a set of data or observations are taken at a specified time in our case which will be the monthly temperature of major cities. The purpose of time series analysis is to extract data from a specified time from the past or a prediction from the future. Which means, with the help of time series analysis it is possible to forecast the future situation, explain past behavior, evaluate the current situation or progress and to plan for the future compiling the results and prediction which in our case will provide the major cities with their temperature changes, paint a picture of the future condition based on the ongoing trend and to sustain a plan on controlling the temperature rise.\\

Time series analysis works with four major components which are trend, seasonality, irregularity and cycle. Trend basically is the common trend observed in a data. In our case, which can be the common trend of temperature growth in certain cities due to global warming, industrialization, pollution etc. Seasonality is how the data changes based on a certain time frame in a visible basis such as during December certain cities can have cooler climate and a much hotter climate in April due to seasonal changes. Irregularities are unexpected changes in the data or graph which can be drastic changes in temperature in certain months due to natural disasters. And lastly the cycle is the general cycle the graph follows \cite{1}.\\

What makes this model so viable for forecast and prediction is that it can work with a single variable. In time series analysis we can work our predictions based on a single “time” variable. It's simpler because in this case the irregularity or any missing data of any other variable or variables don’t hamper the equations. To make it clear, in a simple regression model the equation may look like $y = mx + c$ where the value of “y” is always dependent on the value of “x”. But in time series we can simply work with “y”. If we look at a simple time series equation there, we can put the equation for the value of “y” as in $y_t = y_{t-1},y_{t-2},y_{t-3},…,t$ where the variable y in time t can be traced back using variable y in t-1, t-2 to the furthest variable available with some error adjustment. This means any missing data or any irregularities in data can be adjusted with the help of previous data. The given equation falls under the AR model which is one of the simplest machine learning models. Like this other existing model in time series analysis exists such as AR, MA, ARIMA, Prophet, Nuroprophet etc.\\

\subsection{Seasonality}
Time series characteristic in which data changes on a consistent and predictable basis throughout the year, is called seasonality. Any recurrent or repeated change or trend over the course of a year is said to be seasonal. A seasonal pattern occurs if a time series is impacted by cyclic factors, for example the time throughout the year or weekday. These are the pulsating forces at work, working in a consistent and predictable manner over the course of a year. They follow the same or roughly the same pattern throughout the period of a year. This deviation will be obvious in the time series if the data is collected hourly, daily, weekly, quarterly, or monthly. Seasonality has a set and predictable periodicity \cite{2}. 

\subsection{Trend}
Data pattern that shows how a series progresses over time to comparably greater or lower values, is called trend.  To be specific, as the time series' slope grows or falls, a trend is detected. A trend generally lasts for a short time before evaporating; it isn't the same every time. A trend is what happens when data reveals a long-term increase or decrease. It's not need to be in order. The term "changing direction" refers to when a trend flips from rising to dropping. The trend projects how likely the data is to increase or decrease with time. A trend is a long-term, averaged, smooth pattern. The increase or fall does not need to be in the similar path throughout a set length of time. The propensity may appear to increase, decrease, or remain constant over time. The overall trend, however, must be positive, negative, or stable \cite{3}. \\

\subsection{Stationarity}
Stationarity refers to an attribute of time series which can be observed independent of time. In the case of motionless time series in general, predictable patterns are not observed over time.The series will look horizontal (with some cyclic behavior) on time graphs, with constant variance. If the time series is not stationary, one of the following procedures can be used to make it stationary. Given the series , a new series can be created by differentiating the data\cite{4}. One point less from the actual data will be held on the differenced data. One difference is enough although the difference can be done more times. Also, we can arrange in some curve to the data and then model from that arrangement in case a trend is present. A simple fit, for instance a straight line, is used most of the time since the goal is to lessen the trend. With the help of logarithm or square root of the series we can assist in stabilizing non-constant variance. We can imply a proper constant to make all the data positive before applying the transformation on negative data. Erasing this constant from the model can get us expected values and projections for future points.
 \\

\subsection{Recurrent Neural Networks (RNN)}
When it comes to thinking or decision making, we don’t think from scratch. For instance, during basic communication we don’t trace back to the first word every time to figure out the next suitable word to construct a proper sentence but go on based on the previous word we uttered. This persistency is much needed in effective predictions. In order to achieve that we need recurrent neural networks or RNN. 
RNN is a modified version of traditional neural network where the persistency is maintained as it has a loop. In this pattern it can recur the previous events constantly and work based on that\cite{5}.\\

\begin{figure}[htbp]
\centering
\includegraphics[scale=0.15]{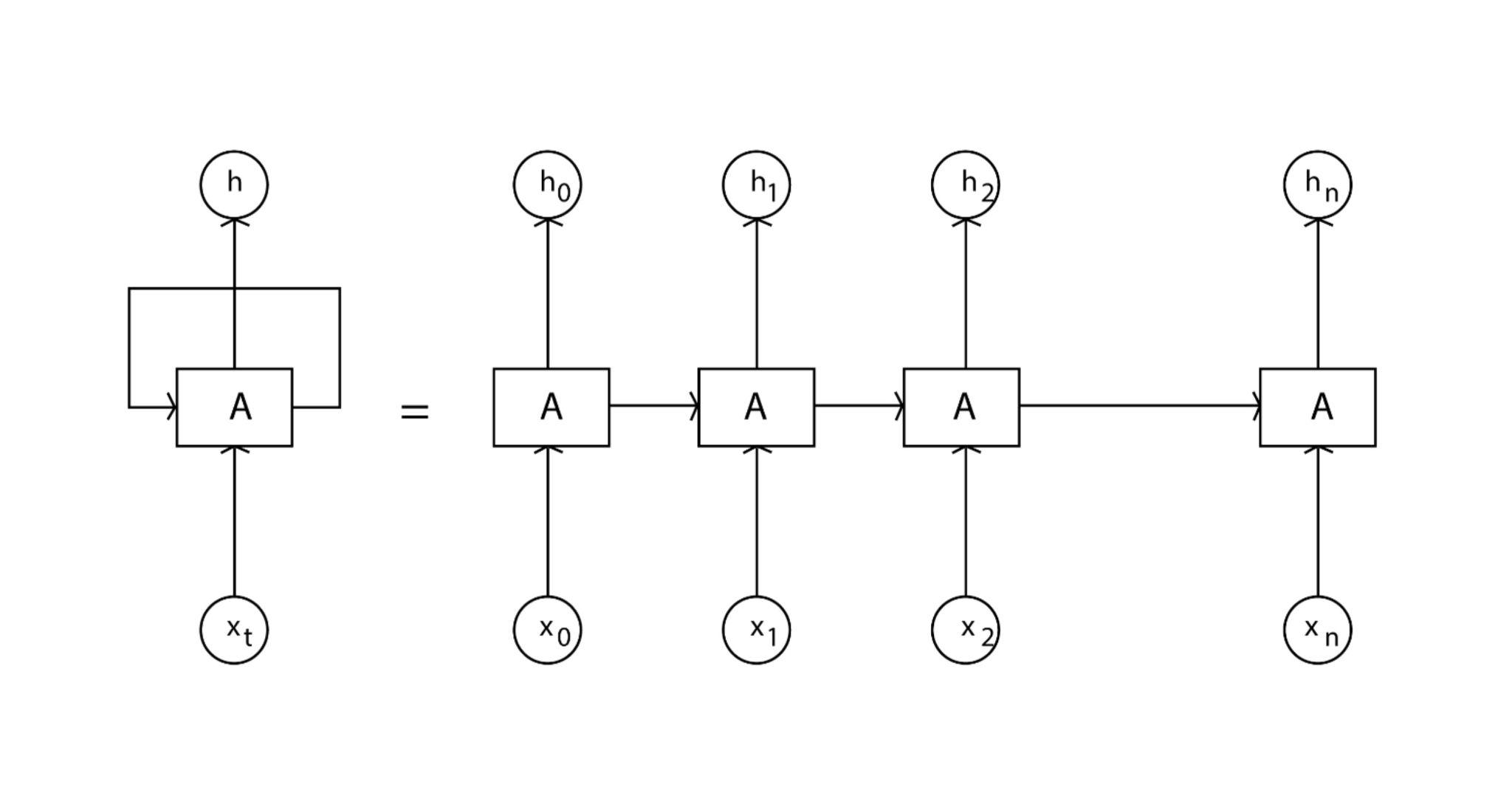}
\caption{Dataflow in RNN}
\label{2.1}
\end{figure} 

To illustrate this with the help of figure \ref{2.1}, here we can see that for a RNN “A” which is getting xt data as input in order to return ht as output, it is creating multiple copies. Each time a new input is generated it is passed down to its successor creating a loop. This chain-like nature reveals that recurrent neural networks are intimately related to sequences and lists. They’re the natural architecture of a neural network to use for such data. \\

\section{Literature review}
Paper\cite{6} works with 4 types of regressors which are Linear regressor, isotonic regressor, support vector regressor and polynomial regressor. All these regressors are used on Monthly Average temperature of Bangladesh Along with Rain data collected during 1901-2015 time period. In preprocessing, the dataset has been split into 3 parts depending on the 3 seasons of Bangladesh, summer, rainy and cold. The 4 regressors are then put into work on the 3 separate data models. the result of the training dataset showed that isotonic regressor has performed better than other regressors. 3 types of errors which are Mean Squared Error, Mean Absolute Error , Median Absolute Error with a R2 score which represented the extent of fluctuation that excluded the autonomous factors in the model helped verifying the result of the training sets. But, when attempted to run the isotonic regressor on the testing datasets, the isotonic regressor failed by giving a constant value for the future temperature of Bangladesh for 2019 to 2040. SVR and Polynomial regressor of degree 3 performed quite well as the findings were shown in the paper. Ashfaq et al. concluded the paper by showing SVR and Polynomial Regressor of Degree 3 can be considered the best for predicting the future average temperature values. \\ 

This study\cite{7}, was done by a few members from the research group, “Kilimanjaro ecosystem under global change: Linking biodiversity, biotic interactions and biogeochemical ecosystem process” with the aim to gather high resolution climate information that are essential for various applications. 14 various types of machine learning algorithms were used here to forecast the monthly air temperature across the Kilimanjaro region. As a result, more accurate results were gathered than what the orthodox kriging approach can gather. Several linear and non-linear models were used the linear models being GLM, GAM, PCR, PLS, SvmLinear and non-linear being avNNet, KNN, NNET, svmRadical alongside cubist, ctree, gbm, rf regression trees. The linear models generally tried to minimize the sum of the square errors either with a focus bias or a variance. Non-linear models provide predictions based on the amplitude of various models used to quantify the distance between the predictor variables and the model's closest known group. The regression trees divided the training dataset into categories based on response values that were comparable. The prediction model is chosen based on rules that are appropriate for the predictors of the variables.\\

The authors of the paper\cite{8}, invested their attention to a model based on EMD(Empirical mode decomposition) and LS-SVM(Least Squares Support Vector Machine). They used a dataset containing the monthly average temperature during 1951-2003. To start, EMD was applied to decompose the time series into a series of various scales of Intrinsic mode function. For these IMFs, the appropriate kernel function and model parameters are used to construct LS-SVM to predict. Based on the input and output objects here they have used linear kernel function and the RBF. The authors also used EMD-LSSVM model to predict the temperature values. And then, the Root mean Square Error(RMSE) and Relative Error (RE) model are used to verify the prediction accuracy. Their research result conveys that among the three models used the EMD-LSSVM models perform better than the separate LS-SVM and the RBF. Here, EMD-LSSVM model predicts with high accuracy and smaller volatility. According to the paper, The reason behind this is that a non-stationary time series can be made a series of stable single components with certain regularity.\\

The authors in the paper \cite{9}, used time series analysis to forecast weather temperature. The paper successfully showed that it’s  possible to predict the evolution of temperature by means of the ARIMA (Auto-Regressive Integrated Moving Average) models.  The research is based on collected data of the past 150 years where the reference period 1850-1899 was more pronounced for Europe and Belgium(MIRA). They used the 10 year moving average in the analysis. They stated that regression analysis would be an inappropriate approach to model the trend of a time series, since it assumes time as an independent variable, whereas, time series are characterized by the dependence of their data. Hence, to analyze dependent data they proposed Arima models. Moreover, they pointed out that ARIMA models have a weak point that the models require  the time series to be stationary before starting the analysis. Therefore, to identify the appropriate ARIMA model for a time series, it’s needed to remove the major seasonality in order to obtain a stationary series. To compare models they used the AIC (Akaike information criterion) as a measure of goodness of fit. Time Series Modeling v4.30 software was used as the main tool for computations. To find whether the series is stationary or not they used the Box-Pierce test.\\

The focus of this paper \cite{10}, is to ensure better understanding of climate change based on different anthropogenic emission scenarios. They have focused on finding how short term emissions as such have a long term effect on climate change and approached machine learning to find this information. Their work is highly data driven and they have proposed building a surrogate climate model using sets of GCM simulations performed in recent years Hadley Centre Global Environment model 3. Taking in account the different senecios the initial sudden response shown in the first few years are considered to be short term and then when the global mean temperature reaches a steady state are considered to be long term in their research. The task consists of taking short term response as “x” and long-term response as “y” and learning the function of x “f(x)”. The mapping to be constructed using Ridge regression, Gaussian Process regression with a linear kernel. Both Ridge and the GPR increases the accuracy by a fair margin but the error is lessened using GPR.\\

In this paper \cite{11}, various weather figure methods were considered and the results of applying different types of machine learning  and ANN algorithms on weather forecasting were also compared. It also explains how meteorologists blend a few techniques like synoptic forecasting, persistence forecasting, computer forecasting and statistical forecasting to forecast weather. Outputs of different types of models like RBF-HPSOGA, RBF-GA, Gaussian SVM, Wavelet SVM and RBF-NN etc. were analyzed. Moreover, authors investigated various information digging approaches for forecasting climate. Finally, RBF NN utilizing Hybrid PSGOA and Wavelet based SVM, gave the maximum execution productivity.\\

This paper \cite{12} mainly talks about solving two problems and comes to a solution. The first problem is to synchronize the problem. In this paper they made a synchronization process to forecast the temperature. The second problem focuses on spatial-equilibration between sites that looks at the relative correlations of primary and proxy variables. But both of their jobs are to forecast the temperature. The forecasted time is not limited. This research is forecasting the future condition of temperature. They used “Oscillation Discovery and Prediction” which was important to forecast the longtime temperature and this is the main thing of this paper.\\

This paper \cite{13} talks about very useful methods. It also talks about regional temperature forecasting and long-term global temperature forecasting. In long-term global temperature forecasting they are shown how they can predict the gt by taking SI, SOD, CO2, Sulfate, ENSO as input. The diagram is given below. RNN method is very useful to predict temperature. And there are also some error calculations which are talked about in this paper. These methods are very useful and it will be very helpful. As there is also talking about those five errors it will make the prediction perfect. They have shown four tables to explain the inputs outputs and the configuration. They wanted to make a prediction which would become more accurate and they have succeeded. They also talked about the Stacked Denoising Auto-Encoders which gives 97.94\% accuracy. On the other hand ANN gives 94.92\% accuracy.\\

In this study report\cite{14}, Researchers suggested a strategy of machine learning in order to forecast the concentration of PM2.5 in a city of Ecuador, Quito. In order to do that, they used the meteorological data of this highly elevated city. To split the data into multiple groups according to the concentrations of PM2.5, machine learning algorithms are utilized. In this specific classification problem, supervised learning approaches such as BTs and  L-SVM are used to develop models. A regression is performed using BT, L-SVM, and NN. Using CGM  to do regression outperforms other commonly used machine learning methods such as NN, LSVM, and BT. Also, using time series algorithms to discover trends over long periods of time should increase the accuracy of the prediction and allow forecasting of the concentrations of PM2.5 for a longer period of time. Moreover, Three fundamental meteorological elements which are direction of wind, wind speed and precipitation are the base of this model, all of which have a direct impact on pollution.\\

In the paper \cite{15}, authors have done a survey on  the latest studies of deep learning-based weather forecasting considering the aspects of the design of NN architectures. Here, the author focuses on deep learning techniques in weather forecasting by comparing the existing DLWP studies. Then they analyzed the pros and cons of  DLWP by comparing it with the conventional NWP, and summarized the potential of  DLWP. They have discussed three types of data and they're Multi-dimensional real-type data, Satellite image data, Long time sequence data. They have also mentioned two types of DNN models. One is basic DNN models such as Autoencoders CNN and LSTM and another one is typical hybrid DNN models. STConvS2S, ConvLSTM, TrajGRU, PredRNN, MetNet models are used for weather state prediction and Hybrid CNN-LSTM, multi-channel convolutional encoder-decoder models are used for extreme weather detection. This work mainly is on extreme event detection on planetary-scale data and again they have inspected the outputs from the climate model’s results to explain the climate changes by the year 2100. \\

In this paper \cite{16}, authors aimed to outperform the traditional methods of weather forecasting by using robust machine learning techniques. Here, they predicted the highest and lowest temperature for seven days using weather data of the past two days. They used a variation of functional regression model and linear regression model. Among these two, the first one was capable of capturing the trends in the weather. Though both models got outperformed by the professional weather forecasting services, the discrepancy between the professional forecasting and their models reduced quickly in the prediction of the later days. They also think that their models might outperform the professional ones in predicting temperature for longer time periods. Moreover, they found out that the  linear regression model outperformed the functional regression model. Nevertheless, they think that  the latter would have performed better if they had based their forecasts on the weather data of four or five days.\\

In this paper \cite{17}, the authors attempted to develop an efficient low cost weather forecasting system using machine learning which would work in remote areas. They used data analytics and machine learning algorithms like random forest classification to predict the condition of the weather. The main target was to predict whether it would rain or not on a particular day depending on the factors like humidity, temperature and pressure.  They found out that the most important factor while predicting rain is humidity followed by temperature and pressure. They have used these machine learning algorithms in python on a Raspberry Pi 3 B board. Other hardwares like the BMP180 pressure sensor and DHT11 humidity and temperature sensor were also a necessary part of this work.\\

In this study \cite{18}, the authors suggested a weather prediction method that leverages historical information from several weather stations to develop basic machine learning models that can provide reasonable predictions for certain atmospheric patterns in the near future in a short amount of time. Furthermore, they argue that using data from many adjacent weather stations rather than data from only the region for which weather forecasting is being done is preferable. Because the expected outcomes are continuous quantitative values, such as temperature, they employed regression techniques. Because it ensembles several decision trees when making decisions, Random Forest Regression (RFR) is demonstrated to be the best regressor. Ridge Regression (Ridge), Support Vector Regression (SVR), Multi-layer Perceptron Regression (MLPR), and Extra-Tree Regression (ETR) are just a few of the regression techniques employed.
\section{Methodology and Implementation}
The intention of this research is to build a model which can predict the temperature of future years in an accurate manner. The dataset which is chosen for the research holds a vast data of previous patterns of rising temperature of many major cities in the globe. We have built our model based on machine learning and deep learning algorithms. Our model is fed the pre-processed data to train the model and is tested with testing data. Time series analysis is a major model that deals with this type of problem. Models like ARIMA, SARIMA and PROPHET have been used to build our model. To utilize the full potential of time series analysis we used LSTM and CNN. \\
To build an accurate model we have planned our work in several steps. Those steps are-

\begin{enumerate}
    \item \textbf{Data preprocessing:} We have ensured all the preprocessing our datasets need in this step. It can prove useful for the accuracy of our model. \\
    
    \item \textbf{Splitting Dataset:} The dataset has been split into a Training set and a testing set in this step.\\
    
    \item \textbf{Research and Develop: } We reviewed previous works which has been done on this specific topic and try to learn from them to build a more accurate and robust model. \\
    
    \item \textbf{Build and Training proposed model:}  At this level, we have built our proposed models and trained it with the aforementioned Training dataset.
 \\
    
    \item \textbf{Testing and measuring the model:}  Testing data was feed to the model to test its accuracy and also compare the accuracy with training accuracy\\
    
    \item \textbf{Visualizing the result:}  The predicted values has been be presented in this final step.\\
\end{enumerate}

\begin{figure}[htbp]
\centering
\includegraphics[scale=0.3]{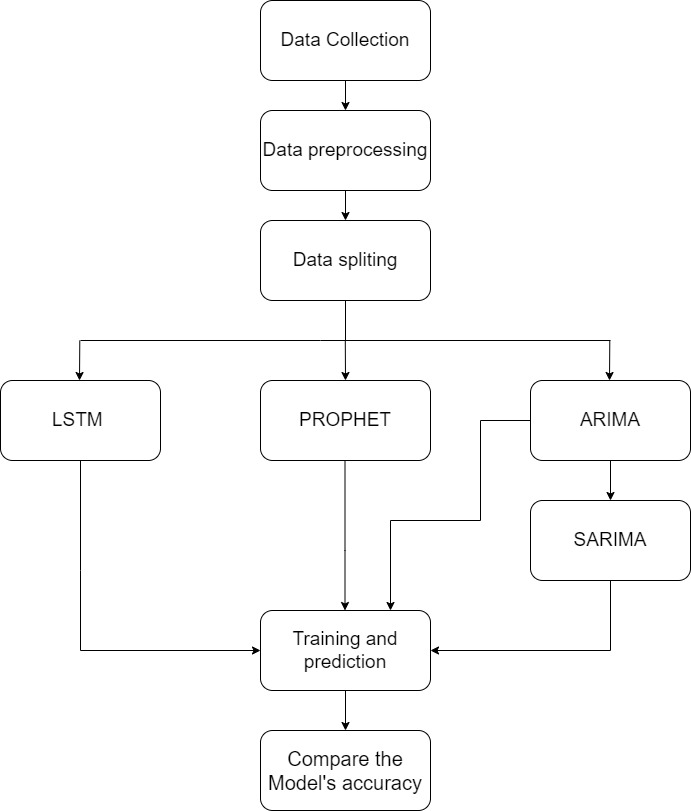}
\caption{flowchart for achieving research objectives}
\label{4.1}
\end{figure} 

\subsection{Dataset}
In order to get appropriate temperature predictions, a very well detailed dataset is a must need. Therefore, during our data collection we used the dataset of Average Daily Temperature Archive provided by University of Dayton \cite{19}. This is a dataset which has the daily temperatures of most of the world’s major cities which includes 167 international cities across the various regions of the globe as well as 157 U.S cities. The current dataset includes daily temperature from January 1, 1995 to May, 2020. The dataset contains a total of 8 columns. Among these, 4 columns which include Region, Country, State and city are used to define the location of the data collection and 3 columns which include Month, Day and Year signifies the date of collecting the temperatures. And the 8th or the final column contains the average temperature of a city on a particular date. In table \ref{t4.1}, a filtered version of the dataset is shown. 

\subsection{Data preprocessing}
In order to enhance the generalizability of our model we need to perform various types of data preprocessing processes. These operations would remove the redundant data as well as it would reorganize some of the columns. The processes are- \\

\begin{enumerate}
    \item \textbf{Imputing the null values: } Here the rows which have value -99 in the average temperature column actually contain null values. Therefore, we would replace the null values meaning the row containing value -99, with the average temperature value from its previous row.\\
    
    \item \textbf{Feature Engineering: } we would extract a date column from the Month, Day and Year column. This would be much more effective than the previous format. Moreover, we would drop the previously existing Month, Day and Year column.\\
    
    \item \textbf{Dropping Unnecessary Columns: } In the selected dataset, all the countries except the U.S, do not have any element in the State column. Hence, we would drop the State column as well as the Region since these information are redundant in regards to our model.\\
    
    \item \textbf{Data Splitting: } The entire data frame was split into 80\% and 20\% for the training set and testing set respectively. Again, 20\% of the training set was used in the validation. For the individual 3 models we have performed data pre-processing accordingly. The batch size we created from the dataframe has a size of 32. Furthermore we have created mini batches which has size of 5 for a smoother training process.
\\
\end{enumerate} 

\begin{table}[!ht]
    \small
    \centering
    \begin{tabular}{|l|l|l|l|l|l|l|l|}
    \hline
        Region & Country & City & Day & Year & AvgTemp \\ \hline
        Africa & Algeria & Algiers & 1 & 1995 & 64.2 \\ \hline
        Africa & Algeria & Algiers & 2 & 1995 & 49.4 \\ \hline
        Africa & Algeria & Algiers & 3 & 1995 & 48.8 \\ \hline
        Africa & Algeria & Algiers & 4 & 1995 & 46.4 \\ \hline
        Africa & Algeria & Algiers & 5 & 1995 & 47.9 \\ \hline
    \end{tabular}
    \caption{Raw data from the dataset}
    \label{t4.1}
\end{table}

\begin{table}[!ht] 
    \large
    \centering
    \begin{tabular}{|l|l|}
    \hline
        Date & AvgTemperature \\ \hline
        1995-01-01 & 64.2 \\ \hline
        1995-01-02 & 49.4 \\ \hline
        1995-01-03 & 48.8 \\ \hline
        1995-01-04 & 46.4 \\ \hline
        1995-01-05 & 47.9 \\ \hline
    \end{tabular}
    \caption{Data after preprocessing}
\end{table}

\subsection{Architecture}
In this section, we have explained our proposed model that we have used to predict the future temperatures. We have selected 3 algorithms for our model which are given below:

\subsubsection{LSTM}
One drawback of RNN is although it can predict from recent events it lacks the concept of context. For instance, during the prediction of temperature there are some exceptions that might occur such as a natural disaster. These events go against the normal state and need to take the context in account. LSTM takes in account the long term dependencies \cite{20}. What modifies an RNN model to a LSTM model is that the repeating module here has four layers. \\
\begin{figure}[htbp]
\centering
\includegraphics[scale=0.15]{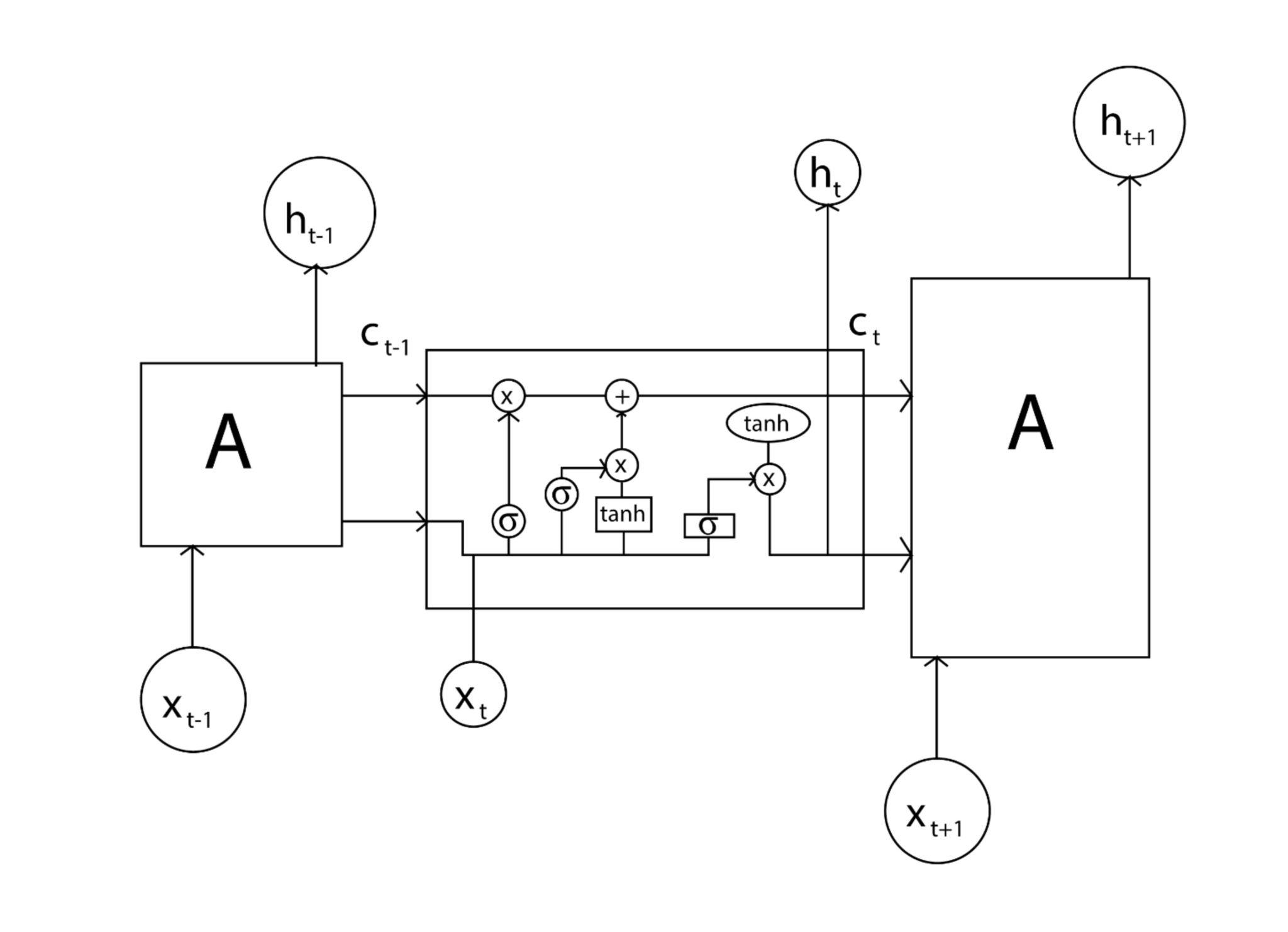}
\caption{Dataflow in LSTM}
\label{6.1}
\end{figure} 

Here in the given figure (\ref{6.1}) we can see the full process of how LSTM operates. For starting we have data from a past event being sent to a successor state. Here the forget gate is receiving the memory of a past event. In another part, a sigmoid layer is called the “forget gate layout” which looks at $h_{t\:-\:1}$  and xt  and outputs a number between 0 and 1 where 1 is to keep the information and 0 is to get rid of the information \cite{21}.

\begin{equation}
f_t\:=\:\sigma\left(W_f\:\cdot \left(h_{t\:-\:1},\:x_t\right)+b_f\right)
\end{equation}                                     
 Afterwards, another sigmoid layer called “input gate layout” decides which values should be updated and a tanh layer creates a vector for new values that can be updated.
                                    \begin{equation}f_t\:=\:\sigma\left(W_f\:\cdot \left(h_{t\:-\:1},\:x_t\right)+b_i\right)\end{equation}
                                    
                                    \begin{equation}C_t \:=\:\tanh \left(W_c\:\left(h_{t\:-\:1},\:x_t\right)+b_i\right)\end{equation}
                                   
Next step is to update the old cell state from $C_{t\:-\:1}$  to $C_{t\:}$. If we multiply the old state by $f_t\:$ and add $i_t\cdot C_t\:$ we get the new candidate value. This value is scaled by how much we want to update each state values. In our code here we drop information of the past events and imply new values.
                                   \begin{equation}
                                       C_{t\:}=\:f_t\:\cdot \:\:C_{t\:-\:1\:}+\:\:i_t\cdot C_t
                                   \end{equation}
Finally the output we get is a filtered version of the cell state. It is done by first running a sigmond layer to decide the parts of the cell state that will be given as outputs. Then the cell state is pushed through tanh and is multiplied by the output of the sigmond gate to filter out the output needed. For the weather forecast it is where it decides weather overcast clouds may lead to rain or not.
                                    \begin{equation}
                                        o_t\:=\:\sigma\left(W_o\:\cdot \left(h_{t\:-\:1},x_t\right)+b_o\right)
                                    \end{equation}
                                    \begin{equation}
                                        h_t=o_t\cdot tanh\left(C_t\right)
                                    \end{equation}

\begin{figure}[htbp]
\centering
\includegraphics[scale=0.5]{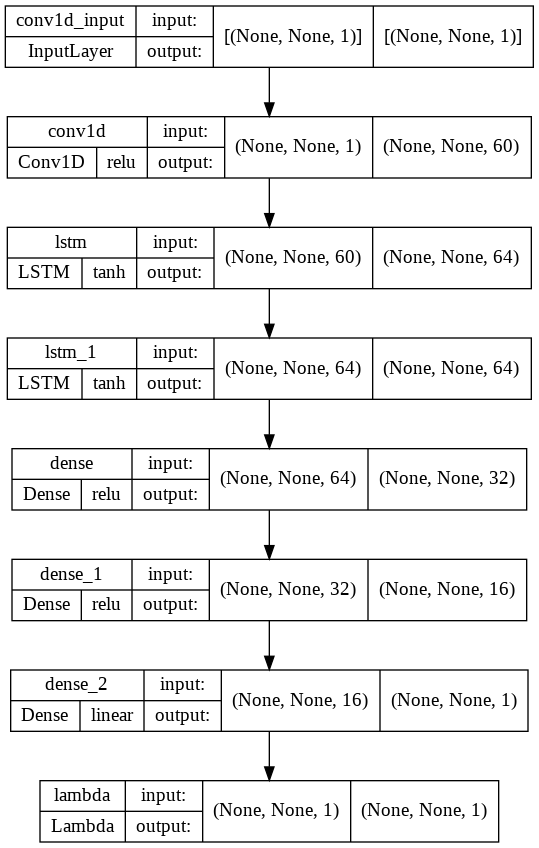}
\caption{Proposed LSTM model}
\label{6.2}
\end{figure}

Our proposed CNN-LSTM model is shown in the Figure \ref{6.2}, we have used keras and tensorflow libraries to build our CNN-LSTM model. The details of this implementation is shown below,

\begin{enumerate}
    \item \textbf{Convolutional layer: } In our model, we have used the conv1D layer in keras. One conv1D is used for the input layer which has a total of 192 parameters.\\
    
     \item \textbf{LSTM layer: } We have decided to use a total of 2 LSTM layers for the model. These 2 layers have tanh as their activation function. Total of 24382 and 33024 parameters are used in these 2 layers of LSTM respectively.\\
     
      \item \textbf{Dense layer: }  After the LSTM layers, we have selected 3 dense layers from keras which are also called the hidden layers. These dense layers have Relu as their activation function and have a total of 2080, 1056 and 33 parameters respectively\\
\end{enumerate}

\newpage
\begin{figure}[htbp]
\centering
\includegraphics[scale=0.35]{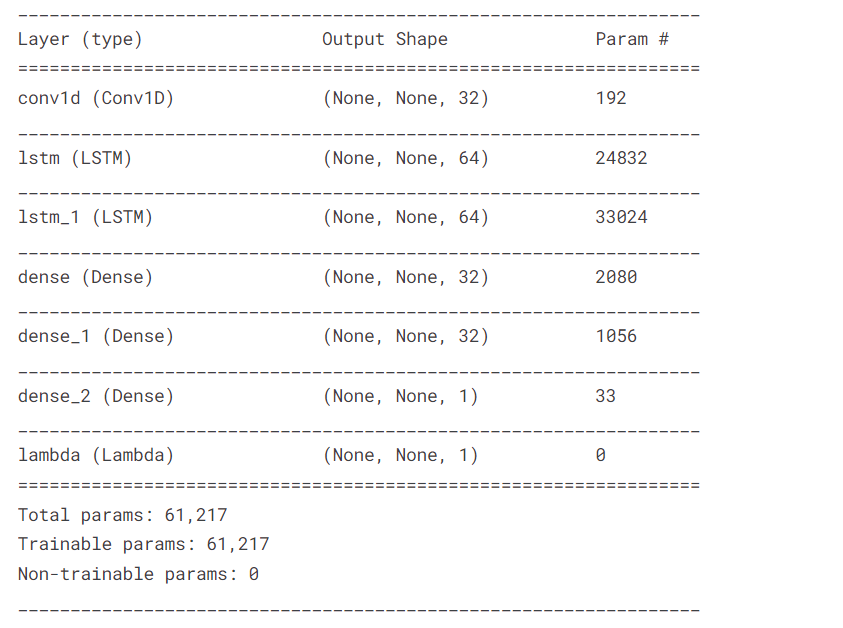}
\caption{LSTM model summary}
\label{6.3}
\end{figure} 

\subsubsection{Prophet}
We have chosen Prophet as our second model to perform the prediction on the selected data frame. Prophet is a time series data forecasting process based on an additive model that fits non-linear trends with yearly, weekly, and daily seasonality, as well as holiday impacts. Prophet was released as open source software by Facebook's Core Data Science team. It works well with time series with strong seasonality and data from multiple seasons of historical data. Prophet decomposes data into three main model components: trend, seasonality and holidays and they are combined in the given equation-
 
                                      \begin{equation}
                                          yt\:=\:g\left(t\right)\:+\:s\left(t\right)\:+\:h\left(t\right)\:+\:et
                                      \end{equation}

Here, g(t) describes a piecewise-linear trend, s(t) describes periodic changes like  various seasonal patterns, h(t) represents holiday effects which take place on irregular schedules over a day or a period of days and $et$ is the error term which represents any individual changes which are not explained by the model \cite{22}. 

\subsubsection{ARIMA}
We know in time series analysis the data has to be stationary. Which means it has to have a content mean over time. But sometimes this requirement is not full-filled. In that case the AR or MA model becomes obsolete and we have to resort to the ARIMA model. ARIMA means Auto Regressive Integrated Moving Average where instead of predicting the time series itself the prediction is done based on one stamp of the series from its previous time stamp. So, for a series "$y_t$" we will take a portion of it "$z_t$" which we can simply define as $z_t = a_{t\:-\:1}\:-\:a_t$ which is the data of some month subtracted by data from previous months. Therefore, even if the graph itself is not stationary we can still divide it into small stationary parts. For our research it will assist us greatly when it comes to the unpredictable changes in value of temperature. ARIMA model consists of three parameters p, d, q where p is the parameter of the AR part d for the integration part and q for the MA part. So, for a simple $ARIMA_{(1,1,1)}$ model the equation is supposed to look like:
                    \begin{equation}
                        Z_{t\:}=\:\phi _{1\:}Z_{t\:-\:1}\:\:+\:\theta _1\:e_{t\:-\:1}\:+e_t
                    \end{equation}
Now to extract values for the main function $y_t$:
            \begin{equation}
               y_t=z_{t-1}a_{t-1}=\:z_{t-1}\:z_{t-2}\:a_{t-2}=\dots =\sum _{i\:-\:1}^{t\:-\:l}z_{t-i}\:+\:a_l
            \end{equation}
Here we have the last data value at al and so that’s our end value of data \cite{23}.

\subsubsection{SARIMA}
The SARIMA model is referred to as the Seasonal ARIMA model. The ARIMA model faces a challenge when it comes to seasonality. And seasonality is something highly visible in temperature forecasting as rise and fall in temperature follows a seasonal pattern. And seasonality is not a stationary data which we need in time series analysis. So, the SARIMA model modifies the existing ARIMA model adding seasonal components. We already know in ARIMA model the parameters are p, d, q where p is the parameter of the AR part d for the integration part and q for the MA part and the model looks like $ARIMA_{(p,q,r)}$. To take seasonality into counts the model is modified to $ARIMA_{p,q,r}P,Q,Rs$ where the capital letters stand for the seasonal parameters of AR terms, differences and MA terms respectively while the value of “s” shows the length of the season. So, in order to forecast wt we should write $wt=yt-yt-l$ where “l” is the length of the dataset. Since SARIMA incorporates both seasonal and non seasonal factors, it can be written as,
                                                            \begin{equation}
                                                                ARIMA(p,d,q) * (P,D,Q)S 
                               \end{equation}
Here, p = non-seasonal AR order, 
d = non-seasonal differencing, 
q = non-seasonal MA order, 
P = seasonal AR order, 
D = seasonal differencing, 
Q = seasonal MA order, 
and S = time span of repeating seasonal pattern.                                                           
We have Used a Sarima model on the temperature data of Rio de janeiro. The model is implemented in the  traditional way. As a regression model, we have performed necessary Data preprocessing to make the training process as smooth as possible. To perform the Sarima model, we had to implement the Arima model on the dataframe as it is shown in the discussion above that the Sarima is actually the Arima with an addition of Seasonal factors\cite{24}. 
\subsection{Implementation}
This section explains the implementation of the proposed model for predicting temperature of the major cities. Python was used for implementation and testing of the proposed model as Python is being used in the majority of machine learning and deep learning models. Tensorflow is used as it is the most popular library for Neural network models. Libraries like Matplotlib, numpy, pandas, fbprophet, seaborn along with sklearn are used to make the model as versatile as possible.
\subsubsection{LSTM Implementation}
To start with, Our proposed model consists of three stages; Data preparation, Training the  model and testing for each algorithm we have used. In the data preparation phase, we have selected Rio De janeiro randomly from our dataset and tested our model corresponding to the city’s temperature. All the data of Rio de Janeiro were separated into a data frame. Then, All the columns are dropped except the Date and AvgTemperature and the false values are removed. In the following figure(\ref{7.1}), the data frame is shown.

\begin{figure}[htbp]
\centering
\includegraphics[scale=0.5]{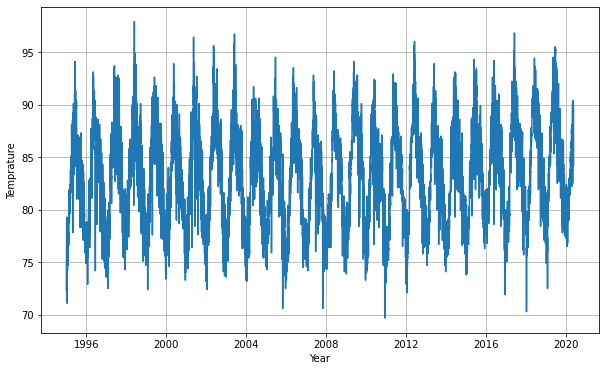}
\caption{DataFrame plotted into graph}
\label{7.1}
\end{figure} 

Here, The total data frame for Rio de Janeiro is plotted on the graph. From 1996 to 2020, Temperature of each day is shown in this graph after dropping all the null and error values.    \\

Again, we have split the dataframe into train and test set. 85 percent of data is split into train set and the rest of 15 percent into test set without shuffling. Then, MinMaxScaler is used to scale the train and test sets. As the most difficult part to separate the data in branches, we have used a keras API to make the branches. 

\begin{figure}[htbp]
\centering
\includegraphics[scale=0.5]{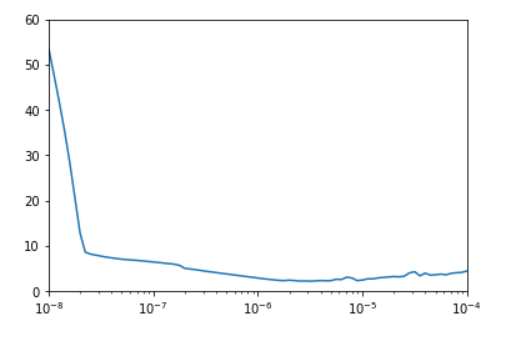}
\caption{Learning Rate of LSTM layer}
\label{7.2}
\end{figure}

After generating the Time series of both the train and test set, we have converted the series into tensor slices. Then, these slices were made into tensor flow windows of size 5. Using the map option we have split the variables into X and Y variables. The batch option helped us to put the variables into mini batches suitable for training. We tried to train our LSTM layer to find the correct learning rate with 100 epochs. By using the LearningRateScheduler we track the correct learning rate. We have taken the step where the learning rate is the steepest shown in the figure (\ref{7.2}).

In the testing phase The model was trained with 500 epochs. Then, we have performed predictions with the test set. 

\subsubsection{Prophet Implementation}
Again, For prophet, we have prepared our data in the same manner as we had done for the LSTM layer. For Prophet, the process was quite straight-forward. The data have been split into the train and test sets. First we fit the train set and test set. Again we  tried to predict the past values. Then, we have compared the values of the actual and predicted value as seen in the following figure (\ref{7.3}).
 
\begin{figure}[htbp]
\centering
\includegraphics[scale=0.35]{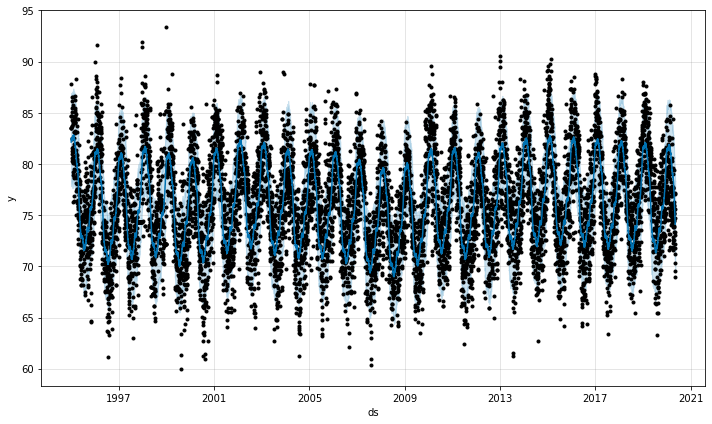}
\caption{The past actual values and the predicted values compared}
\label{7.3}
\end{figure} 

In the figure(\ref{7.3}), the black points are the actual values and the blue lines are the predicted values with trend by the Prophet model. The Temperature is on the y axis plotted against the years which are plotted in the x axis. 

\begin{figure}[htbp]
\centering
\includegraphics[scale=0.3]{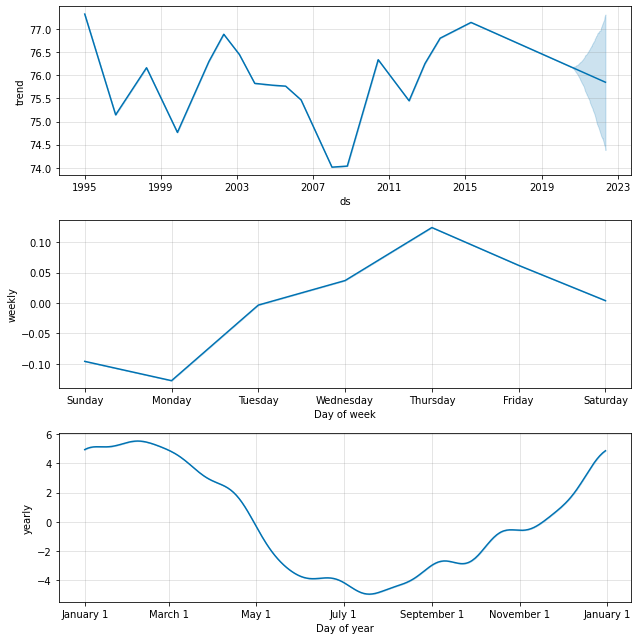}
\caption{Trends of Data in daily, weekly and yearly manner}
\label{7.4}
\end{figure} 
In figure (\ref{7.4}), the trends of the data is plotted using the plot components method from the fbprophet library. The trends of weeks and years are plotted in the y axis against the corresponding dates plotted in the x axis

\subsubsection{ARIMA \& SARIMA Implementation}
\begin{figure}[htbp]
\centering
\includegraphics[scale=0.7]{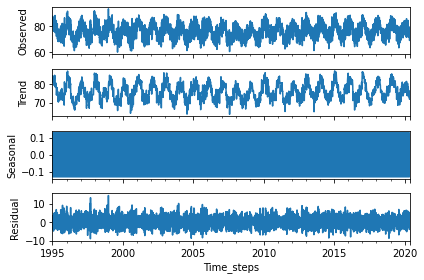}
\caption{Trend and seasonality}
\label{7.5}
\end{figure}

Lastly, we have run the SARIMA model. Here, in the start we have imported the necessary libraries which are numpy, matplotlib, pandas, scikit learn and statesmodel. After importing the data and performing necessary preprocessing, using the decompose method from statesmodel we have plotted the observed, trend, seasonal and residual of the data frame which is shown in figure (\ref{7.5}).

\begin{figure}[htbp]
\centering
\includegraphics[scale=0.7]{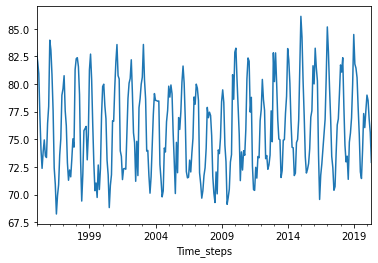}
\caption{Plot of average temperature of months}
\label{7.6}
\end{figure}

The adfuller method from the statsmodel is used to observe if our data is stationary or not. As we have seen that our data is quite stationary, with the P value lower than 0.05 we found out the average temperature of every month of the selected timeframe seen in figure (\ref{7.6}).

\begin{figure}[htbp]
\centering
\includegraphics[scale=0.7]{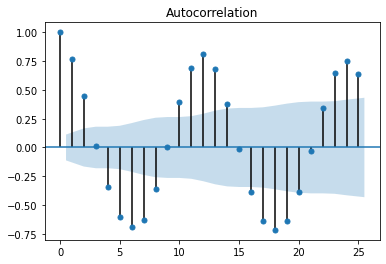}
\caption{Autocorrelation performance}
\label{7.7}
\end{figure}

\begin{figure}[htbp]
\centering
\includegraphics[scale=0.7]{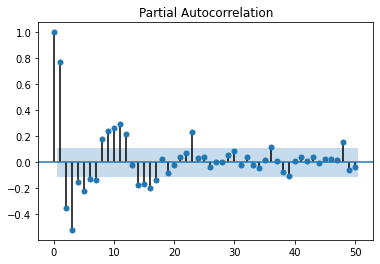}
\caption{Partial Autocorrelation performance}
\label{7.8}
\end{figure}

Now, We have split data into training and test sets where 80 percent of data have gone into the training set and the rest of 20 percent into Test set. Then, we have plotted the autocorrelation (figure \ref{7.7}) and partial autocorrelation (figure \ref{7.8}) of the data frame. Auto correlation and partial autocorrelation are the factors to find out the p, d and q value for implementing the ARIMA model.To check the stationarity we have used the autocorrelation function and the partial autocorrelation. We have also implemented a custom way to find the correct value for p,d and q which is a for loop to train the model with all the different combinations of p,d and q ranging 0-8. After finding out the best combination for p,q and d which is (7,0,2) by checking the least RMSE value given by our method. We trained our arima model with the following parameters and predicted using the test Data. As We have determined the parameters of AR, I and MA according to the behavior between the ACF and PACF plots in the figure \ref{7.6} and the custom method we used, we tried to train our SARIMA model with the S value of 12. In the end we plotted the predicted results from the SARIMA model.

\section{Results}

\subsection{Predictions according to LSTM}
As we trained our LSTM model for 500 epochs using our batch data, We have predicted the temperature of Rio de janeiro for the test data. From the figure, we can see the predicted data from 2017 to 2020 in the following figure (\ref{8.1}).

\begin{figure}[htbp]
\centering
\includegraphics[scale=0.5]{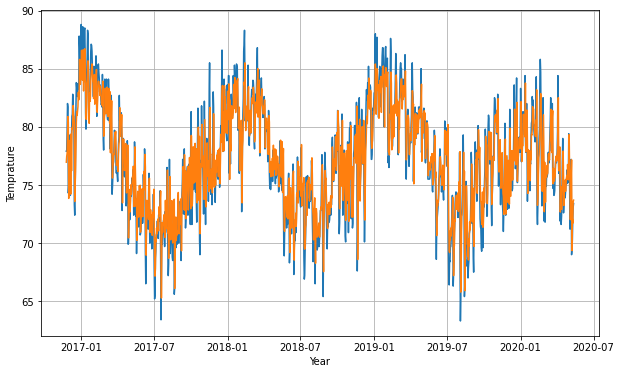}
\caption{Predictions performed by LSTM}
\label{8.1}
\end{figure}
\begin{figure}[htbp]
\centering
\includegraphics[scale=.5]{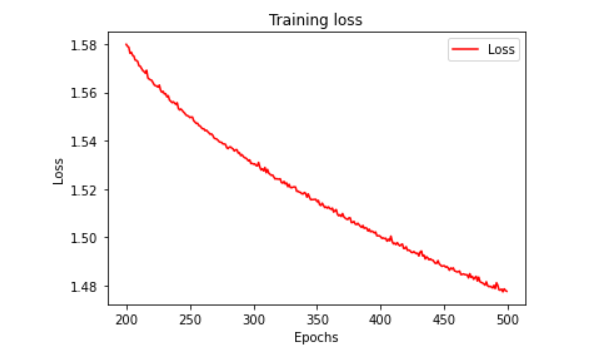}
\caption{Zoomed training loss for LSTM}
\label{8.3}
\end{figure}

 
It can be seen from the graph of figure \ref{8.1}, the predicted values are quite close to the actual values. The orange lines represent the predicted values while the blue ones represent the actual values for the year from 2017 to 2020. After that the training loss is plotted against Epochs in x axis. It can be hard to understand how the Training loss has decreased over Epochs. For that a zoomed view of Training losses are shown in the figure \ref{8.3}. In the zoomed view it is shown that the training loss has decreased from 1.58 to 1.48 as epochs increased from 200 to 500. \\

\begin{figure}[htbp]
\centering
\includegraphics[scale=0.7]{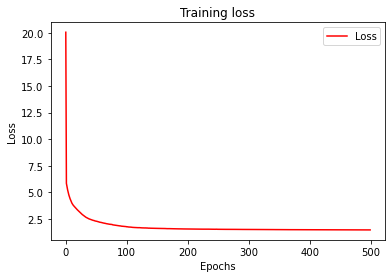}
\caption{Training loss compared to Epochs for LSTM}
\label{8.2}
\end{figure}

\newpage
Predicted values from the LSTM model have been enlisted in the table (\ref{t8.1}) with a comparison between the actual values of the last set of dates from the test set. If we examine the values of both actual and predicted, it is quite eminent that the predictions are quite good in terms of actual values. For example, on the date 2020-05-09 the predicted and actual value are almost the same. Again for the next day at the table, A similar case has taken place.

\begin{table}[!ht]
    \centering
    \begin{tabular}{|l|l|l|}
    \hline
            Date & actual & Predicted  \\ \hline
        2020-05-03 & 71.2 & 75.361137  \\ \hline
        2020-05-04 & 71.8 & 72.810982  \\ \hline
        2020-05-05 & 74.7 & 72.864136  \\ \hline
        2020-05-06 & 77.2 & 75.708511  \\ \hline
        2020-05-07 & 69.0 & 76.833656  \\ \hline
        2020-05-08 & 69.5 & 69.785004  \\ \hline
        2020-05-09 & 70.3 & 70.931351  \\ \hline
        2020-05-10 & 72.2 & 72.706345  \\ \hline
        2020-05-11 & 73.1 & 72.295151  \\ \hline
        2020-05-12 & 73.4 & 73.690948  \\ \hline
    \end{tabular}
    \caption{Prediction according to LSTM}
    \label{t8.1}
\end{table}

\subsection{Predictions using Prophet}
\begin{figure}[htbp]
\centering
\includegraphics[scale=0.3]{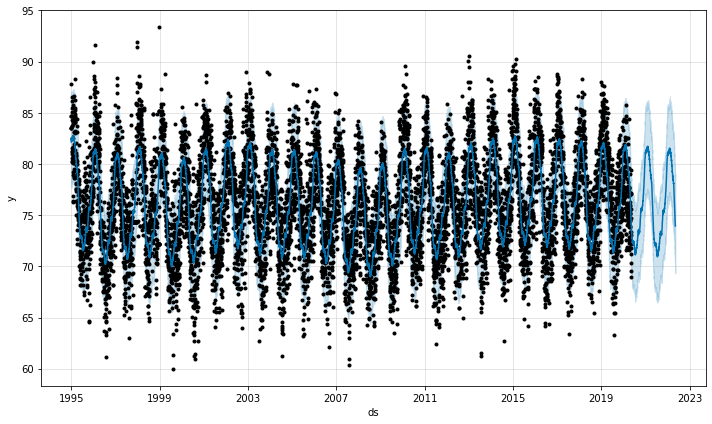}
\caption{Prediction for 2020 to 2022 of  Rio de Janeiro}
\label{8.4}
\end{figure}

For predictions, Firstly a future dataframe of 730 days is created using the make\_future\_dataframe method from the Fbprophet library. A new set of predictions for dates starting from june 2020 to 2022 june can be seen as the singular blue lines in the Figure (\ref{8.4}). The black dots represent the actual values of the data frame.\\

A comparison between the Actual and predicted values are shown in the following table (\ref{t8.2}). \\

\begin{table}[!ht]
    \centering
    \begin{tabular}{|l|l|l|}
    \hline
        Date & Actual & Predicted  \\ \hline
        2020-05-08 & 69.5 & 74.74  \\ \hline
        2020-05-09 & 70.3 & 74.54  \\ \hline
        2020-05-10 & 72.2 & 74.29  \\ \hline
        2020-05-11 & 73.1 & 74.12  \\ \hline
        2020-05-12 & 73.4 & 74.12  \\ \hline
    \end{tabular}
    \caption{Prediction according to Prophet}
    \label{t8.2}
\end{table}
 Here, it can be seen that the temperature prediction is kind of odd as the predicted temperature has been similar in a relatively short period of time.

\subsection{Predictions from ARIMA}
\begin{figure}[htbp]
\centering
\includegraphics[scale=0.5]{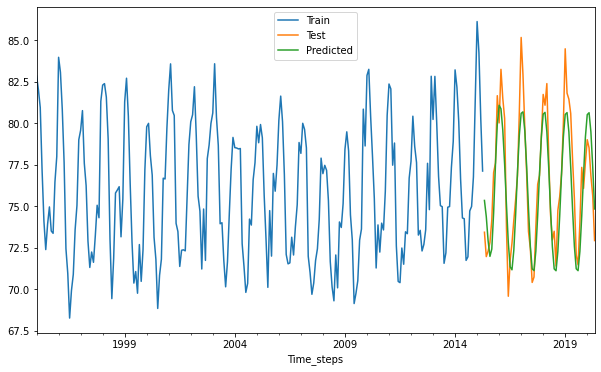}
\caption{Prediction of Arima model}
\label{8.5}
\end{figure}

Before Running the SARIMA, we have performed training and predicting the data with ARIMA model. As our data was stationary, ARIMA performed quite well predicting the temperature for the test set shown in the figure (\ref{8.5}). We have also shown the values of  prediction from month 05 to month 09 of 2015 (table \ref{t8.3}).

\begin{table}[!ht]
    \centering
    \begin{tabular}{|l|l|l|}
    \hline
        DATE & PRED\_TEMPERATURE & ACTUAL\_TEMP  \\ \hline
        2015-05-31 & 75.342545 & 73.425806  \\ \hline
        2015-06-30 & 74.367738 & 71.956667  \\ \hline
        2015-07-31 & 72.908608 & 72.293548  \\ \hline
        2015-08-31 & 71.970500 & 72.812903  \\ \hline
        2015-09-30 & 72.409934 & 74.150000  \\ \hline
    \end{tabular}
    \caption{Prediction according to Arima}
    \label{t8.3}
\end{table}

\subsection{Predictions of SARIMA}
\begin{figure}[htbp]
 \centering
 \includegraphics[scale=0.5]{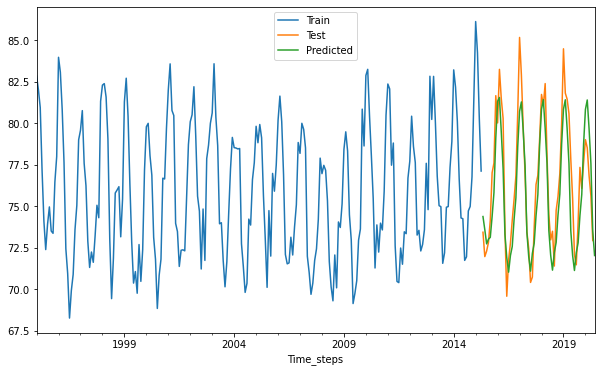}
 \caption{Prediction from Sarima}
 \label{8.6}
 \end{figure}

\begin{figure}[htbp]
 \centering
 \includegraphics[scale=0.4]{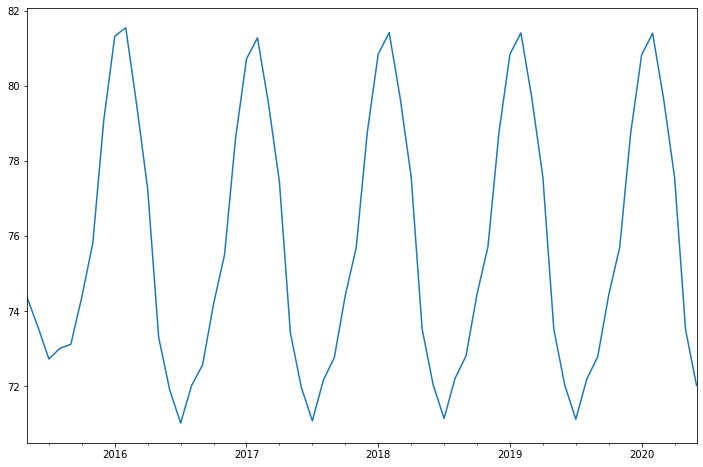}
 \caption{Prediction from Sarima}
 \label{8.7}
 \end{figure}

Finally As we have run the SARIMA model which has given us much better results than the upper 2 algorithms, it can be seen that the SARIMA parameters are well fitted and the predicted values are following the actual values (table \ref{t8.4} and figure \ref{8.6}) and also the seasonal pattern (figure \ref{8.7}).

 \begin{table}[!ht]
    \centering
    \begin{tabular}{|l|l|l|}
    \hline
        DATE & ACTUAL & PREDICTED  \\ \hline
        2015-05-31 & 73.425806 & 74.371810  \\ \hline
        2015-06-30 & 71.95667 & 73.580960  \\ \hline
        2015-07-31 & 72.293548 & 72.724876  \\ \hline
        2015-08-31 & 72.812903 & 73.008618  \\ \hline
        2015-09-30 & 74.150000 & 73.117321  \\ \hline
    \end{tabular}
    \caption{Prediction according to Sarima}
    \label{t8.4}
\end{table}

 Now, As we analyze the error values for each case in the following table (\ref{t8.5}), it is quite clear that SARIMA is performing better than the three algorithms we have used here. The prophet model has performed in the poorest manner while the LSTM model has performed quite well but fell short to SARIMA and ARIMA.

\begin{table}[!ht]
    \centering
    \begin{tabular}{|l|l|l|l|}
    \hline
        Model & MAE & MSE & RMSE  \\ \hline
        LSTM & 1.7365966 & 5.367847 & 2.31686  \\ \hline
        Prophet & 2.913137 & 13.62539 & 3.691260  \\ \hline
        ARIMA & 1.506151 & 3.365973 & 1.834659  \\ \hline
        SARIMA & 1.274921 & 2.672121 & 1.634662  \\ \hline
    \end{tabular}
    \caption{MAE, MSE and RMSE score on Rio de Janeiro}
    \label{t8.5}
\end{table}

As we have observed that SARIMA is performing better on the Rio de Janeiro time frame. We decided to perform Sarima in a different city, Data. We chose Delhi, which is one of the most polluted cities in the world and tested our model on the temperature of Delhi. Here, unlike the temperature of Rio de janerio, the data of delhi was not stationary as we decompose the data and tested with the adfuller method from statsmodel library. we had to change the hyperparameters of ARIMA and SARIMA according to the ACF and PACF of the temperature. We chose the best fit values for p, d and q which are 5,0 and 5 and trained the model. 

 \begin{figure}[htbp]
 \centering
 \includegraphics[scale=0.5]{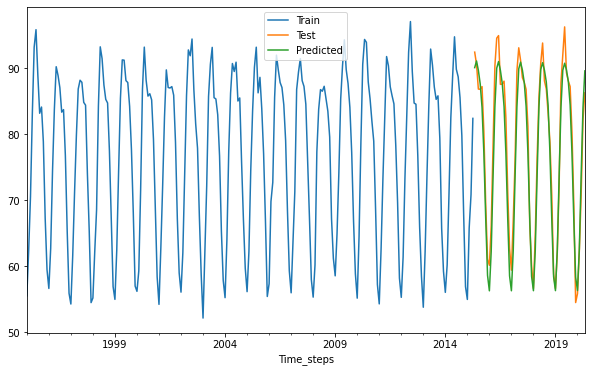}
 \caption{Prediction from Arima on Delhi}
 \label{8.8}
 \end{figure}

 \begin{figure}[htbp]
 \centering
 \includegraphics[scale=0.35]{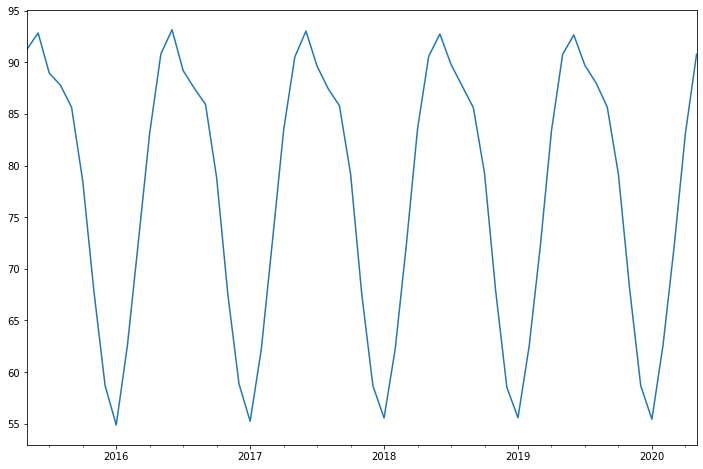}
 \caption{Zoomed prediction from SARIMA}
 \label{8.9}
 \end{figure}

 \begin{figure}[htbp]
 \centering
 \includegraphics[scale=0.5]{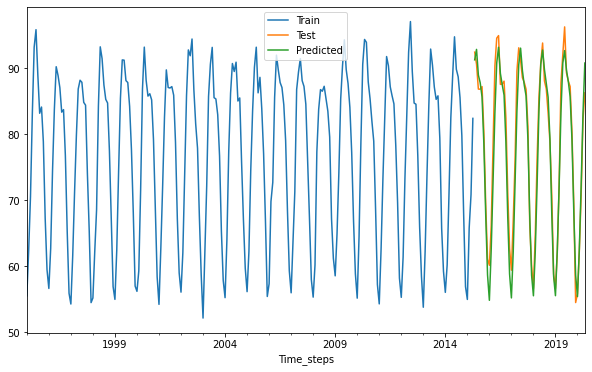}
 \caption{Prediction according to SARIMA on Delhi}
 \label{8.10}
 \end{figure}
 
 Predictions of the Sarima model on Delhi are shown in the aforementioned figure (\ref{8.10}). The predictions which are the green line are following the orange lines which represent the test set values. The predicted values of table (\ref{t8.6}) also represent the comparison with the test values.
 
 \begin{table}[!ht]
    \centering
    \begin{tabular}{|l|l|l|}
    \hline
        Date & Actual & Predicted  \\ \hline
        2015-05-31 & 92.429032 & 91.265368  \\ \hline
        2015-06-30 & 90.6 & 92.844741  \\ \hline
        2015-07-31 & 86.835484 & 88.965473  \\ \hline
        2015-08-31 & 86.819355 & 87.791113  \\ \hline
        2015-09-30 & 87.213333 & 85.648202  \\ \hline
    \end{tabular}
    \caption{Comparison between predicted and actual temperature}
    \label{t8.6}
\end{table}

The mae, mse and rmse values of the Sarima model on the Delhi dataframe is shown here in the table (\ref{t8.8}). 
\begin{table}[!ht]
    \centering
    \begin{tabular}{|l|l|l|l|}
    \hline
        ~ & MAE & MSE & RMSE  \\ \hline
        ARIMA & 2.448460 & 8.512704 & 2.917654  \\ \hline
        SARIMA & 2.178788 & 7.419787 & 2.723928  \\ \hline
    \end{tabular}
    \caption{MAE,MSE and RMSE of ARIMA and SARIMA on Delhi}
    \label{t8.8}
\end{table}

Lastly, we have created a future data frame from the 6th month of 2022 from the 7th month of 2023. We have tried to predict this future data frame with the sarima model which is seen in the table (\ref{t8.7}).

\begin{table}[!ht]
    \centering
    \begin{tabular}{|l|l|}
    \hline
        Date & Predicted values \\ \hline
        2022-06-30 & 92.885653  \\ \hline
        2022-07-31 & 89.781013 \\ \hline
        2022-08-31 & 87.966480  \\ \hline
        2022-09-30 & 86.049364  \\ \hline
        2022-10-31 & 79.336561  \\ \hline
        2022-11-30 & 68.093446  \\ \hline
        2022-12-31 & 59.117201  \\ \hline
        2023-01-31 & 55.784209  \\ \hline
        2023-02-28  & 62.455916 \\ \hline
        2023-03-31  & 72.261380  \\ \hline
        2023-04-30  & 83.093126  \\ \hline
        2023-05-31  & 90.382022  \\ \hline
        2023-06-30  & 92.758995  \\ \hline
        2023-07-31  & 89.879544  \\ \hline
    \end{tabular}
    \caption{Future prediction by sarima model based on Delhi}
    \label{t8.7}
\end{table}
 
 \subsubsection{Conclusion}
 Changes in temperature have caused great concerns in the recent decades. The orthodox methods of calculating and predicting this increase of temperature around the world are becoming more and more obsolete. So, in order to get better insight and accurate calculation modern approaches such as using Machine learning algorithms are necessary for accuracy and efficiency. As we implied Time series analysis using LSTM in different models such as Prophet, ARIMA and SARIMA and put them into comparison we came to the conclusion that the best performing model for the prediction has been SARIMA. Implying this model in Rio De Janeiro we found that when applied to the Prophet or ARIMA the error value is greater than what occurs when SARIMA is implied.
Afterwards the model was used in a different city which is Delhi this time as it has a higher pollution level. This time though when using the adfuller method from the statsmodel library the data came out not to be stationary. So, basically it not only helped us to predict the temperature of Delhi it also portrayed the imbalance that pollution creates in temperature.

\subsection{Future work}
Although we have just portrayed this way of predicting temperature we do have some future aspirations. Recently we have found A global GHG inventory from 1990-2017 on International Greenhouse Gas Emissions which will see us compare how much pollution effect the trend of the temperature. Furthermore, we’ve found a dataset on Carbon Dioxide emission rates. We will try to merge these two or at least one dataset with the dataset we used in our paper. For these we will need some time to further work on it. Nevertheless, we are looking forward to develop our own model or exploring other new methods of ensemble learning to get more accurate results.  So we have some future aspirations to work on this project.

\printbibliography

\end{document}